\documentclass[11pt]{article}

\usepackage[final]{acl}

\usepackage{times}
\usepackage{latexsym}
\usepackage[T1]{fontenc}
\usepackage[utf8]{inputenc}
\usepackage{microtype}
\usepackage{inconsolata}
\usepackage{float}

\usepackage{graphicx}
\usepackage{booktabs}
\usepackage{amsmath}
\usepackage{url}
\usepackage{pgfplots}
\pgfplotsset{compat=1.18}

\title{From Global Benchmarks to Local Evaluations:\\ Benchmarking LLMs for the German Public Sector}

\author{
  \textbf{Camilla Dalerci$^*$},
  \textbf{Thilo Michael$^*$},
  \textbf{Robin Schaefer$^*$},
  \textbf{Daniel Weinland$^*$}
\\
\\
Innovations Department, Bundesdruckerei GmbH, Berlin, Germany,
\\
\texttt{firstname.lastname@bdr.de}
}

\begin{document}
\maketitle
\def\thefootnote{*}\footnotetext{These authors contributed equally to this work.}\def\thefootnote{\arabic{footnote}}
\def\thefootnote{}\footnotetext{This work will be presented at the Eval4SD workshop, co-located with KONVENS 2026 in Hamburg, Germany.}\def\thefootnote{\arabic{footnote}}

\begin{abstract}
Public institutions face a persistent challenge in selecting LLMs suited to their specific context. 
Existing benchmarks, however, are of limited use as they primarily reflect English-language and US-centric settings, and often only evaluate task performance.
In this paper, we present first results of MÖVE, a holistic evaluation framework for the German public sector, examining three rarely considered governance dimensions: energy consumption, provider transparency, and knowledge of German-party positions. 
Our results reveal significant trade-offs, with no single model excelling across all dimensions: estimated energy consumption varies more than 60-fold and is not explained by model size alone, information disclosure varies systematically across providers, and European models do not exhibit stronger knowledge of German party positions. 
Model selection for public institutions thus cannot rely on performance rankings alone. 
Instead, evaluations should also reflect the governance requirements of the deployment context.
\end{abstract}

\section{Introduction}
\label{sec:introduction}

Large language models (LLMs) are increasingly used in high-stakes environments, including public administration. 
At the same time, model selection is no longer merely a choice between one or two well-known providers. 
Public institutions can now choose from a rapidly expanding and heterogeneous landscape of models that differ in size, performance, energy requirements, documentation practices, and contextual knowledge. 

As LLM-generated text increasingly informs policy analysis and public-sector decision-making, the need for robust and locally grounded benchmarking practices becomes more pressing \cite{reuel-etal-2024-betterbench}. 
General benchmark performance alone offers limited evidence about whether a model is suitable for a particular institutional, linguistic, or societal context. 
The ability to identify potential harms and anticipate failures within intended use-case scenarios has therefore made model evaluation an essential discipline \cite{chang-etal-2024-survey}. 

Established LLM benchmarks predominantly measure task performance on English-language datasets and frequently reflect US- or UK-specific contexts \cite{wang-etal-2019-superglue,hendrycks_measuring_2021,majithia-etal-2026-citizenquery}. 
Multilingual benchmarks extend coverage to additional languages \cite{adelani-etal-2025-irokobench,susanto-etal-2025-sea} but often rely on translated versions of existing datasets \cite{singh-etal-2025-global}, thereby failing to address the domain or context gap. 
Moreover, benchmarks generally prioritize task performance over the broader governance concerns that a more holistic evaluation would address \cite{liang-etal-2023-holistic}. 
Thus, while a model may perform well on conventional benchmarks, it may, e.g., consume disproportionate resources or provide insufficient documentation, which are crucial criteria for the broad adoption of LLMs. 
These limitations are particularly consequential for public institutions, as they must not only assess whether a model can be applied to a task, but also whether its deployment can be justified to decision-makers, oversight bodies, and the public. 

In this paper, we present first results of MÖVE (\textit{Modelle für die öffentliche Verwaltung evaluieren}), a holistic evaluation framework for the German public sector.\footnote{MÖVE is conceived as a \textit{living} benchmark, which will be regularly updated both with respect to the assessed models and the evaluation criteria. Our leaderboard can be found here: \url{https://moeve.bundesdruckerei.de/} (in German). Our framework code can be found here: \url{https://github.com/Bundesdruckerei-GmbH/moeve-lmbench}. For more details on the evaluation criteria, datasets, prompts and results, see our MÖVE framework paper: \citet{dalerci-etal-2026-moeve}.}
While the frameworks' current status includes seven performance and governance criteria, in this paper we focus on the evaluation of 39 LLMs across three governance dimensions.
We estimate inference-time \textit{energy consumption} using task-specific output statistics, assess \textit{provider transparency} through a manually verified matrix of 21 questions across seven documentation domains, and evaluate \textit{political knowledge} using 4,788 official positions from 64 German political parties. 
While requiring different evaluation methods, the three dimensions address a common question: whether a model is suitable for deployment in a specific public-sector context beyond its performance on a generic task. 

\section{Benchmark Design}

\subsection{Scope and Target Groups}

We evaluate 39 open-weight and proprietary LLMs\footnote{See Appendix \ref{app:transparency_scores} for a list of evaluated models.} from 13 providers across three governance dimensions relevant to public-sector deployment: inference-time energy consumption, provider transparency, and \textit{knowledge} of German political-party positions.
The benchmark was developed through a stakeholder-informed process involving semi-structured feedback sessions with public-sector representatives and exchanges with practitioners.
Following recommendations that benchmark design should be grounded in intended uses and stakeholder needs \cite{reuel-etal-2024-betterbench}, we distinguish four target groups: 1) AI decision-makers, 2) public-administration domain experts, 3) IT departments and security-critical institutions, and 4) broader civil society.
Their respective needs concern strategic model comparison, operational suitability, secure and compliant deployment, and public accountability.

\begin{table*}[ht]
  \centering
  \begin{tabular}{lccc}
    \toprule
    Dataset & Task & Type & Size \\
    \midrule
    Eur-Lex-Sum \cite{aumiller2022eur} & Summarization & Existing & 850 \\
    Swiss Leading Decision Summarization \cite{rasiah2023scale} & Summarization & Existing &  1,530 \\
    KIKC Summary & Summarization & Private & 40 \\
    German Ministry Publications (Summaries) & Summarization & Private & 1,530 \\
    German-QuAD \cite{moller-etal-2021-germanquad} & QA & Existing & 4,710 \\
    KIKC QA & QA & Private & 72 \\
    FAQ Law & QA & Private &  129 \\
    KIKC Topics & Topic Extraction & Private & 205\\
    German Ministry Publications (Topics) & Topic Extraction & Private & 4,199 \\
    \bottomrule
  \end{tabular}
    \caption{Overview of datasets used for calculating energy consumption during model evaluation. \emph{Type} indicates whether a dataset is pre-existing or specifically constructed for the benchmark. The latter are not released to the public. Dataset size is reported with respect to the evaluation unit used in each task, e.g., \emph{summaries} in summarization tasks.}
    \label{tab:dataset_overview}
\end{table*}

\subsection{Energy Consumption}

Energy consumption can accumulate substantially when models are deployed at scale, making energy efficiency an operational as well as an environmental consideration \cite{strubell-etal-2019-energy,wu-etal-2025-unveiling}.
To enable comparison between locally deployed and API-accessed models, we estimate inference-time energy consumption using EcoLogits \cite{rince-banse-2025-ecologits}. 
The framework estimates energy use from the model’s total parameter count, active parameter count for mixture-of-experts architectures, and number of generated output tokens. 
Parameter information for open-weight models was collected from official documentation; for proprietary models without disclosed parameter counts, we relied on the assumptions provided by EcoLogits. 

Rather than applying a fixed output length, we use the actual output-token counts recorded during model evaluation across nine German-language datasets (Table~\ref{tab:dataset_overview}): four summarization datasets, three question-answering datasets, and two topic-extraction datasets.
We report mean estimated energy consumption in watt-hours per inference request, both by task and across tasks.

\subsection{Provider Transparency}

Meaningful risk assessment depends on provider transparency. 
Without adequate information about training data, bias mitigation, computational resources, and model limitations, adopting institutions cannot independently evaluate the risks transferred to them.
This information is also increasingly relevant in the European regulatory context, particularly in light of the documentation requirements for providers of general-purpose AI models established under Article 53 of the EU AI Act\footnote{\url{https://eur-lex.europa.eu/eli/reg/2024/1689/oj/eng}.}. 

Provider transparency is evaluated using a structured dataset of 21 questions across seven domains: model identification; architecture and properties; distribution and access; use and deployment; training and data; computational resources and energy consumption; endorsement of the EU General-Purpose AI Code of Practice.
The questions were derived from the Code’s Transparency Chapter and Model Documentation Form\footnote{\url{https://digital-strategy.ec.europa.eu/en/policies/contents-code-gpai}.}, which operationalize documentation obligations under Article 53. 

For each model, we collected publicly verifiable information from three source types: official provider websites, official model cards, and technical publications issued by the provider.
Each question was scored as 0 when no information was available, 1 when information was partial, and 2 when clear and verifiable information was provided.
The Code of Practice signature was treated as a binary criterion.
The documentation was first assessed manually, primarily by one researcher, with a subset independently scored by a second.
To support quality assurance, we additionally implemented an automated scoring agent that retrieves the same sources and scores all criteria independently.
The automated and manual assessments initially diverged on 27.4\% of items.
These cases were reviewed and corrected where necessary, and recurring disagreements were encoded as explicit scoring notes, reducing divergence to 16.2\% --- a measure of the final protocol's internal consistency rather than of accuracy, since both the notes and the manual scores were revised in the process.
A second researcher reviewed the complete output.
Across 39 models, the resulting dataset contains 819 model-question assessments.
The full matrix, including the scoring notes, the justification for each score, and the underlying sources, is publicly available.\footnote{\url{https://moeve.bundesdruckerei.de/transparenz}.}

\begin{figure*}[htbp]
    \centering
    \includegraphics[width=\linewidth]{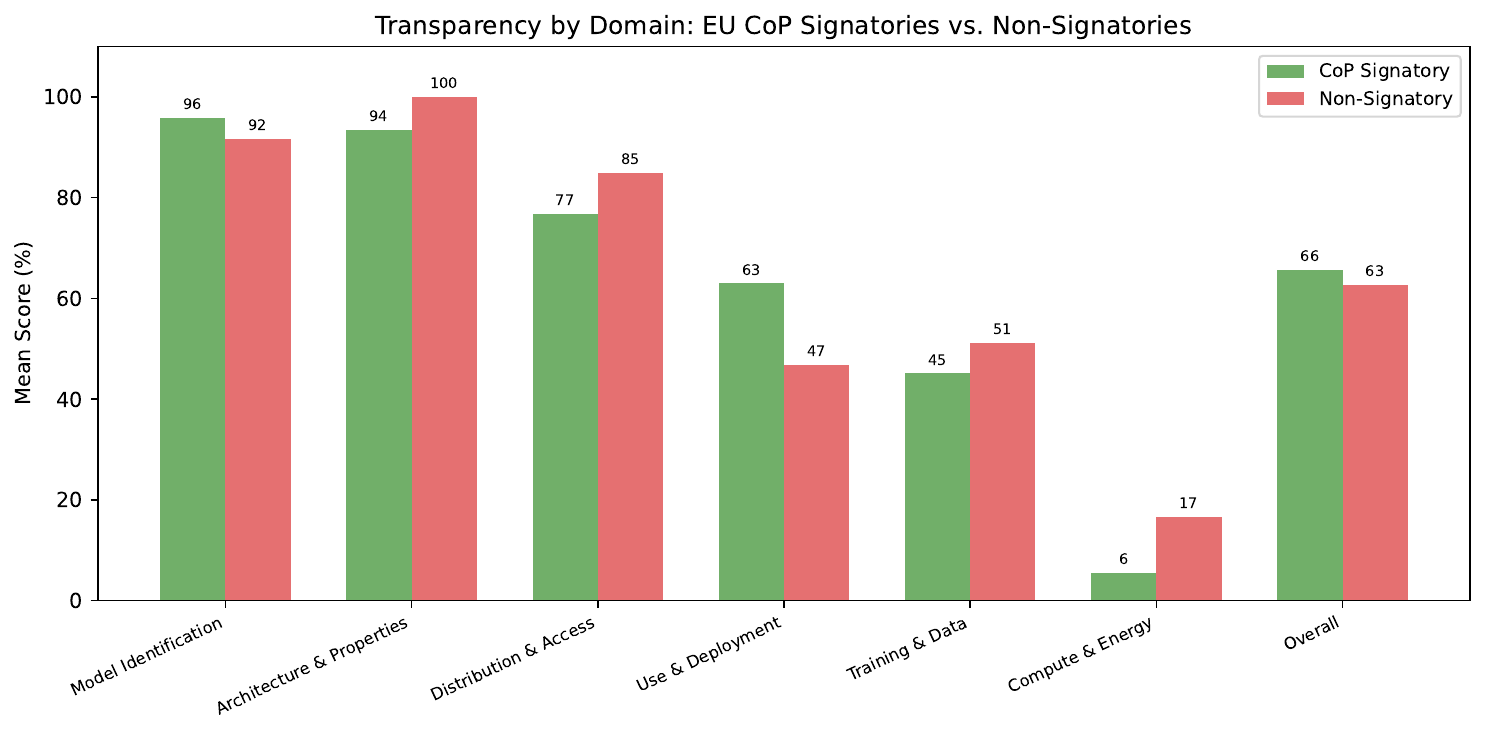}
    \caption{Mean transparency score (\%) by domain for EU~AI~Act Code of Practice signatories vs.\ non-signatories.}
    \label{fig:transparency_signatory_domains}
\end{figure*}

\subsection{Political Knowledge}

Models used in the public sector are required to produce factually accurate outputs. 
\textit{Political knowledge} is a particularly important example because incorrect representations of policy positions may affect citizen-facing information services and policy-related applications.
Existing studies have frequently examined the political preferences attributed to LLMs through instruments such as the Political Compass\footnote{\url{https://www.politicalcompass.org/}.} or national voting-advice applications. 
However, such results are sensitive to prompt formulation, language, and the assumptions used to attribute a political position to a model \cite{rottger-etal-2024-political,helwe-etal-2025-navigating,rettenberger-etal-2025-assessing}. 
We therefore distinguish political knowledge from political bias. 
Rather than measuring political bias in model outputs, we evaluate whether such outputs correctly reflect the publicly documented positions of German political parties. 
This formulation treats \textit{political positioning} as a factual classification task grounded in an external reference rather than as an attribution of partisan preferences to the model. 

\textit{Political knowledge} is evaluated using official party responses from the German Wahl-O-Mat, a voting-advice application maintained by the Federal Agency for Civic Education\footnote{\url{https://www.bpb.de/themen/wahl-o-mat/}.}. 
The dataset covers the four most recent federal elections, i.e., 2013, 2017, 2021, and 2025, and includes all participating parties.
Propositions for which a party did not provide a position were removed, resulting in 4,788 labeled positions (\textit{agree}, \textit{disagree}, or \textit{neutral}) from 64 political parties. 47.5\% of positions agree with the given proposition, while 39.7\% and 12.8\% exhibit a negative or neutral stance, respectively.

Models receive a political proposition, a party, the election year, and the response options in a German-language prompt and are tasked to predict how the specified party positioned itself. 
Each model is queried once per datapoint. 
Performance is measured using classification accuracy against the official party responses. 
Although this task arguably poses a challenge for LLMs, given that party positions may shift over time, we consider it a compelling use case for identifying the upper bounds of LLM performance in a political context.

\section{Results}

Our results reveal substantial variation and trade-offs. 
Estimated mean energy consumption differs by a factor of 63 across the evaluated models, with smaller models generally consuming less energy, although size does not fully determine efficiency. 
Provider documentation is strongest for basic model identification and architectural properties, but remains particularly weak regarding computational resources, energy use, training-data safeguards, and bias-mitigation practices. 
Classification of party positions is challenging across the model set: the highest mean accuracy is 0.671, and model size is positively associated with performance without being sufficient to explain it.
Across the three criteria, no single model achieves the strongest result in every dimension. 

\paragraph{Energy Consumption.}

Estimated energy consumption varies by a factor of 63 across the 39 models, from 0.647 Wh to 40.6 Wh per query, with a median of 1.788 Wh. This estimate should be interpreted with caution for proprietary models, where parameter counts are not disclosed.
Energy requirements also depend strongly on the task: summarization is more energy-intensive than question answering and topic extraction because it produces longer outputs.
Reasoning models incur particularly high energy costs, while several smaller models remain competitive at substantially lower consumption. 
Our analysis therefore indicates that marginal improvements in performance can entail disproportionately large energy costs. 

\paragraph{Provider Transparency.}

Overall transparency scores range from 18 to 38 out of 42, demonstrating substantial differences between models and providers.
Documentation is generally strong for model identification, architecture, and access, but weak for upstream information.
\textit{Compute and energy} is the least transparent domain, with a mean score of 11.5\%; 84.6\% of models disclose no training-energy consumption and 79.5\% provide no measurement methodology.
Training-data safeguards and bias-mitigation practices are also poorly documented (see Appendix \ref{app:transparency_scores} for full results). 
Providers that signed the EU General-Purpose AI Code of Practice score overall only marginally higher than non-signatories, and their advantage is concentrated in downstream information such as intended use and deployment guidance rather than training data, compute, or energy disclosures (Figure \ref{fig:transparency_signatory_domains}). 

\paragraph{Political Knowledge.}

No model produces consistently accurate outputs with respect to German political-party positions (Table \ref{tab:party_positions_top10}).
The two highest-performing models achieve an accuracy of only 0.671 across 4,788 positions from 64 parties.
Although larger models tend to perform better on average, model size does not determine performance: a small GPT-4o Mini matches a large DeepSeek R1 at the top of the ranking, while medium-sized and fine-tuned models also achieve competitive results. 
The highest-performing models span European, US, and Chinese providers, thus providing no evidence that geographical proximity alone predicts higher output accuracy.
Despite substantial variance in model performance, all top-performing models surpass the majority baseline of 0.475.

\begin{table*}[htbp]
    \centering
    \begin{tabular}{cllllc}
        \toprule
        Rank & Model & Organisation & Region & Size & Score \\
        \midrule
        1 & DeepSeek R1 & DeepSeek & China & Large & 0.671 \\
        1 & GPT-4o Mini & OpenAI & USA & Small & 0.671 \\
        3 & GPT-OSS 120B  & OpenAI & USA & Large & 0.667 \\
        4 & Llama 3.3 70B & Meta & USA & Large & 0.663 \\
        5 & Mistral Small 3.1 & Mistral & Europe & Medium & 0.645\\
        6 & GPT-4o	& OpenAI	 & USA & Large & 0.624 \\
        7 & GPT-OSS 20B	 & OpenAI  & USA & Medium & 0.619\\
        8 & Mistral Large 3 & Mistral & Europe & Large & 0.610 \\
        9 & SauerkrautLM Mixtral 8x7B & VAGOSolutions  & Europe & Medium & 0.607 \\
        10 & DeepSeek R1 32B & DeepSeek  & China & Medium & 0.601 \\
        \midrule
        & Majority Baseline & & & & 0.475 \\
        \bottomrule
    \end{tabular}
    \caption{Top-10 models for party-position classification, ranked by mean accuracy. All top-performing models surpass the majority baseline.}
    \label{tab:party_positions_top10}
\end{table*}

\section{Discussion}

The findings of this paper show that model selection for public administration should extend beyond conventional performance rankings.
In particular, the 63-fold variation in estimated energy consumption demonstrates that resource efficiency is not a secondary consideration. 
Models with similar task performance may impose substantially different operational and environmental costs.
Energy consumption should therefore be assessed, especially for high-volume public-sector applications. 

The transparency assessment reveals a persistent accountability gap. 
Providers commonly document basic model properties, access conditions, and intended uses, but disclose notably less about training data, bias mitigation, computational resources, and energy consumption. 
This limits the ability of public institutions to independently assess risks and make evidence-based procurement decisions.

Finally, the political-knowledge results also caution against using a model provider’s geographical origin as a proxy for local suitability. 
Models from European, US, and Chinese providers achieve competitive results on German political-party positions, and no regional group consistently dominates. 
Foreign-developed models should therefore not be excluded based on assumptions about represented contextual knowledge.
At the same time, the modest maximum accuracy shows that all models require evaluation on locally relevant data before deployment in politically sensitive use cases.  

\paragraph{Conclusion.} Our results emphasize the need for a multidimensional, locally grounded approach to LLM evaluation.
Model size, provider reputation, and general benchmark performance alone are insufficient proxies for deployment suitability.
Public administrations should compare models based on specific governance and domain requirements. 
Meanwhile, providers and regulators should strengthen transparency requirements to enable such assessments. 
Overall, these findings suggest that we should move away from universal model rankings and towards evaluations that reflect the deployment context's requirements.

\section*{Limitations}

This study is limited to three evaluation dimensions and a fixed set of models. Future work will extend the benchmark with newer models and additional criteria to provide a more comprehensive assessment of public sector suitability. 

Energy results are comparative estimates rather than complete measurements and for proprietary models rely on EcoLogits' assumptions about undisclosed parameter counts. 
Transparency scores reflect publicly available documentation collected and verified between March and May 2026.

The datasets capture selected aspects of the German public-sector context and cannot represent the full diversity of administrative tasks, political knowledge, or deployment conditions. 

\section*{Acknowledgments}

We thank the anonymous reviewers for their helpful feedback.
In the preparation of this paper, generative AI was used in a supporting capacity for stylistic revision.

\bibliography{custom}

@article{aumiller2022eur,
  title={EUR-lex-sum: A multi-and cross-lingual dataset for long-form summarization in the legal domain},
  author={Aumiller, Dennis and Chouhan, Ashish and Gertz, Michael},
  journal={arXiv preprint arXiv:2210.13448},
  year={2022}
}

@article{chang-etal-2024-survey,
author = {Chang, Yupeng and Wang, Xu and Wang, Jindong and Wu, Yuan and Yang, Linyi and Zhu, Kaijie and Chen, Hao and Yi, Xiaoyuan and Wang, Cunxiang and Wang, Yidong and Ye, Wei and Zhang, Yue and Chang, Yi and Yu, Philip S. and Yang, Qiang and Xie, Xing},
title = {{A Survey on Evaluation of Large Language Models}},
year = {2024},
issue_date = {June 2024},
publisher = {Association for Computing Machinery},
address = {New York, NY, USA},
volume = {15},
number = {3},
issn = {2157-6904},
url = {https://doi.org/10.1145/3641289},
doi = {10.1145/3641289},
journal = {ACM Trans. Intell. Syst. Technol.},
month = mar,
articleno = {39},
numpages = {45}
}

@misc{dalerci-etal-2026-moeve,
      title={{M\"OVE: A Holistic LLM Benchmark for the German Public Sector}}, 
      author={Camilla Dalerci and Thilo Michael and Robin Schaefer and Daniel Weinland},
      year={2026},
      eprint={2606.13111},
      archivePrefix={arXiv},
      primaryClass={cs.CL},
      url={https://arxiv.org/abs/2606.13111}, 
}

@inproceedings{helwe-etal-2025-navigating,
    title = "{Navigating the Political Compass: Evaluating Multilingual {LLM}s across Languages and Nationalities}",
    author = "Helwe, Chadi  and
      Balalau, Oana  and
      Ceolin, Davide",
    editor = "Che, Wanxiang  and
      Nabende, Joyce  and
      Shutova, Ekaterina  and
      Pilehvar, Mohammad Taher",
    booktitle = "Findings of the Association for Computational Linguistics: ACL 2025",
    month = jul,
    year = "2025",
    address = "Vienna, Austria",
    publisher = "Association for Computational Linguistics",
    url = "https://aclanthology.org/2025.findings-acl.883/",
    doi = "10.18653/v1/2025.findings-acl.883",
    pages = "17179--17204",
    ISBN = "979-8-89176-256-5"
}

@misc{hendrycks_measuring_2021,
    title = {Measuring {Massive} {Multitask} {Language} {Understanding}},
    url = {http://arxiv.org/abs/2009.03300},
    doi = {10.48550/arXiv.2009.03300},
    urldate = {2025-11-21},
    publisher = {arXiv},
    author = {Hendrycks, Dan and Burns, Collin and Basart, Steven and Zou, Andy and Mazeika, Mantas and Song, Dawn and Steinhardt, Jacob},
    month = jan,
    year = {2021},
    note = {arXiv:2009.03300 [cs]},
}

@inproceedings{adelani-etal-2025-irokobench,
    title = "{{I}roko{B}ench: A New Benchmark for {A}frican Languages in the Age of Large Language Models}",
    author = "Adelani, David Ifeoluwa  and
      Ojo, Jessica  and
      Azime, Israel Abebe  and
      Zhuang, Jian Yun  and
      Alabi, Jesujoba Oluwadara  and
      He, Xuanli  and
      Ochieng, Millicent  and
      Hooker, Sara  and
      Bukula, Andiswa  and
      Lee, En-Shiun Annie  and
      Chukwuneke, Chiamaka Ijeoma  and
      Buzaaba, Happy  and
      Sibanda, Blessing Kudzaishe  and
      Kalipe, Godson Koffi  and
      Mukiibi, Jonathan  and
      Kabongo Kabenamualu, Salomon  and
      Yuehgoh, Foutse  and
      Setaka, Mmasibidi  and
      Ndolela, Lolwethu  and
      Odu, Nkiruka  and
      Mabuya, Rooweither  and
      Osei, Salomey  and
      Muhammad, Shamsuddeen Hassan  and
      Samb, Sokhar  and
      Guge, Tadesse Kebede  and
      Sherman, Tombekai Vangoni  and
      Stenetorp, Pontus",
    editor = "Chiruzzo, Luis  and
      Ritter, Alan  and
      Wang, Lu",
    booktitle = "Proceedings of the 2025 Conference of the Nations of the Americas Chapter of the Association for Computational Linguistics: Human Language Technologies (Volume 1: Long Papers)",
    month = apr,
    year = "2025",
    address = "Albuquerque, New Mexico",
    publisher = "Association for Computational Linguistics",
    url = "https://aclanthology.org/2025.naacl-long.139/",
    doi = "10.18653/v1/2025.naacl-long.139",
    pages = "2732--2757",
    ISBN = "979-8-89176-189-6"
}

@misc{liang-etal-2023-holistic,
      title={{Holistic Evaluation of Language Models}}, 
      author={Percy Liang and Rishi Bommasani and Tony Lee and Dimitris Tsipras and Dilara Soylu and Michihiro Yasunaga and Yian Zhang and Deepak Narayanan and Yuhuai Wu and Ananya Kumar and Benjamin Newman and Binhang Yuan and Bobby Yan and Ce Zhang and Christian Cosgrove and Christopher D. Manning and Christopher Ré and Diana Acosta-Navas and Drew A. Hudson and Eric Zelikman and Esin Durmus and Faisal Ladhak and Frieda Rong and Hongyu Ren and Huaxiu Yao and Jue Wang and Keshav Santhanam and Laurel Orr and Lucia Zheng and Mert Yuksekgonul and Mirac Suzgun and Nathan Kim and Neel Guha and Niladri Chatterji and Omar Khattab and Peter Henderson and Qian Huang and Ryan Chi and Sang Michael Xie and Shibani Santurkar and Surya Ganguli and Tatsunori Hashimoto and Thomas Icard and Tianyi Zhang and Vishrav Chaudhary and William Wang and Xuechen Li and Yifan Mai and Yuhui Zhang and Yuta Koreeda},
      year={2023},
      eprint={2211.09110},
      archivePrefix={arXiv},
      primaryClass={cs.CL},
      url={https://arxiv.org/abs/2211.09110}, 
}

@misc{majithia-etal-2026-citizenquery,
      title={{The CitizenQuery Benchmark: A Novel Dataset and Evaluation Pipeline for Measuring LLM Performance in Citizen Query Tasks}}, 
      author={Neil Majithia and Rajat Shinde and Zo Chapman and Prajun Trital and Jordan Decker and Manil Maskey and Elena Simperl and Nigel Shadbolt},
      year={2026},
      eprint={2602.04064},
      archivePrefix={arXiv},
      primaryClass={cs.CY},
      url={https://arxiv.org/abs/2602.04064}, 
}

@inproceedings{moller-etal-2021-germanquad,
    title = "{{G}erman{Q}u{AD} and {G}erman{DPR}: Improving Non-{E}nglish Question Answering and Passage Retrieval}",
    author = {M{\"o}ller, Timo  and
      Risch, Julian  and
      Pietsch, Malte},
    editor = "Fisch, Adam  and
      Talmor, Alon  and
      Chen, Danqi  and
      Choi, Eunsol  and
      Seo, Minjoon  and
      Lewis, Patrick  and
      Jia, Robin  and
      Min, Sewon",
    booktitle = "Proceedings of the 3rd Workshop on Machine Reading for Question Answering",
    month = nov,
    year = "2021",
    address = "Punta Cana, Dominican Republic",
    publisher = "Association for Computational Linguistics",
    url = "https://aclanthology.org/2021.mrqa-1.4/",
    doi = "10.18653/v1/2021.mrqa-1.4",
    pages = "42--50"
}

@misc{rasiah2023scale,
      title={{SCALE: Scaling up the Complexity for Advanced Language Model Evaluation}},
      author={Vishvaksenan Rasiah and Ronja Stern and Veton Matoshi and Matthias Stürmer and Ilias Chalkidis and Daniel E. Ho and Joel Niklaus},
      year={2023},
      eprint={2306.09237},
      archivePrefix={arXiv},
      primaryClass={cs.CL}
}

@Article{rettenberger-etal-2025-assessing,
  author   = {Rettenberger, Luca and Reischl, Markus and Schutera, Mark},
  journal  = {Journal of Computational Social Science},
  title    = {{Assessing political bias in large language models}},
  year     = {2025},
  issn     = {2432-2725},
  number   = {2},
  pages    = {42},
  volume   = {8},
  doi      = {10.1007/s42001-025-00376-w},
  url      = {https://doi.org/10.1007/s42001-025-00376-w},
}

@inproceedings{reuel-etal-2024-betterbench,
author = {Reuel, Anka and Hardy, Amelia and Smith, Chandler and Lamparth, Max and Hardy, Malcolm and Kochenderfer, Mykel J.},
title = {{BetterBench: assessing AI benchmarks, uncovering issues, and establishing best practices}},
year = {2024},
isbn = {9798331314385},
publisher = {Curran Associates Inc.},
address = {Red Hook, NY, USA},
booktitle = {Proceedings of the 38th International Conference on Neural Information Processing Systems},
articleno = {685},
numpages = {51},
location = {Vancouver, BC, Canada},
series = {NIPS '24}
}

@article{rince-banse-2025-ecologits,
    title = {{EcoLogits: Evaluating the Environmental Impacts of Generative AI}},
    author = {Rinc{\'e}, Samuel and Banse, Adrien},
    journal = {Journal of Open Source Software},
    volume = {10},
    number = {111},
    pages = {7471},
    year = {2025},
    doi = {10.21105/joss.07471},
    url = {https://joss.theoj.org/papers/10.21105/joss.07471},
}

@inproceedings{rottger-etal-2024-political,
    title = "{Political Compass or Spinning Arrow? Towards More Meaningful Evaluations for Values and Opinions in Large Language Models}",
    author = {R{\"o}ttger, Paul  and
      Hofmann, Valentin  and
      Pyatkin, Valentina  and
      Hinck, Musashi  and
      Kirk, Hannah  and
      Schuetze, Hinrich  and
      Hovy, Dirk},
    editor = "Ku, Lun-Wei  and
      Martins, Andre  and
      Srikumar, Vivek",
    booktitle = "Proceedings of the 62nd Annual Meeting of the Association for Computational Linguistics (Volume 1: Long Papers)",
    month = aug,
    year = "2024",
    address = "Bangkok, Thailand",
    publisher = "Association for Computational Linguistics",
    url = "https://aclanthology.org/2024.acl-long.816/",
    doi = "10.18653/v1/2024.acl-long.816",
    pages = "15295--15311"
}

@InProceedings{singh-etal-2025-global,
  author    = {Singh, Shivalika and others},
  author-full = {Singh, Shivalika and Romanou, Angelika and Fourrier, Cl{\'e}mentine and Adelani, David Ifeoluwa and Ngui, Jian Gang and Vila-Suero, Daniel and Limkonchotiwat, Peerat and Marchisio, Kelly and Leong, Wei Qi and Susanto, Yosephine and Ng, Raymond and Longpre, Shayne and Ruder, Sebastian and Ko, Wei-Yin and Bosselut, Antoine and Oh, Alice and Martins, Andre and Choshen, Leshem and Ippolito, Daphne and Ferrante, Enzo and Fadaee, Marzieh and Ermis, Beyza and Hooker, Sara},
  booktitle = {Proceedings of the 63rd Annual Meeting of the Association for Computational Linguistics (Volume 1: Long Papers)},
  title     = {{Global {MMLU}: Understanding and Addressing Cultural and Linguistic Biases in Multilingual Evaluation}},
  year      = {2025},
  address   = {Vienna, Austria},
  editor    = {Che, Wanxiang and Nabende, Joyce and Shutova, Ekaterina and Pilehvar, Mohammad Taher},
  month     = jul,
  pages     = {18761--18799},
  publisher = {Association for Computational Linguistics},
  doi       = {10.18653/v1/2025.acl-long.919},
  isbn      = {979-8-89176-251-0},
  url       = {https://aclanthology.org/2025.acl-long.919/},
}

@inproceedings{strubell-etal-2019-energy,
    title = "{Energy and Policy Considerations for Deep Learning in {NLP}}",
    author = "Strubell, Emma  and
      Ganesh, Ananya  and
      McCallum, Andrew",
    editor = "Korhonen, Anna  and
      Traum, David  and
      M{\`a}rquez, Llu{\'i}s",
    booktitle = "Proceedings of the 57th Annual Meeting of the Association for Computational Linguistics",
    month = jul,
    year = "2019",
    address = "Florence, Italy",
    publisher = "Association for Computational Linguistics",
    url = "https://aclanthology.org/P19-1355/",
    doi = "10.18653/v1/P19-1355",
    pages = "3645--3650"
}

@inproceedings{susanto-etal-2025-sea,
    title = "{{SEA}-{HELM}: {S}outheast {A}sian Holistic Evaluation of Language Models}",
    author = "Susanto, Yosephine  and
      Hulagadri, Adithya Venkatadri  and
      Montalan, Jann Railey  and
      Ngui, Jian Gang  and
      Yong, Xianbin  and
      Leong, Wei Qi  and
      Rengarajan, Hamsawardhini  and
      Limkonchotiwat, Peerat  and
      Mai, Yifan  and
      Tjhi, William Chandra",
    editor = "Che, Wanxiang  and
      Nabende, Joyce  and
      Shutova, Ekaterina  and
      Pilehvar, Mohammad Taher",
    booktitle = "Findings of the Association for Computational Linguistics: ACL 2025",
    month = jul,
    year = "2025",
    address = "Vienna, Austria",
    publisher = "Association for Computational Linguistics",
    url = "https://aclanthology.org/2025.findings-acl.636/",
    doi = "10.18653/v1/2025.findings-acl.636",
    pages = "12308--12336",
    ISBN = "979-8-89176-256-5"
}

@inbook{wang-etal-2019-superglue,
  author    = {Wang, Alex and Pruksachatkun, Yada and Nangia, Nikita and Singh, Amanpreet and Michael, Julian and Hill, Felix and Levy, Omer and Bowman, Samuel R.},
  publisher = {Curran Associates Inc.},
  title     = {{SuperGLUE: a stickier benchmark for general-purpose language understanding systems}},
  year      = {2019},
  address   = {Red Hook, NY, USA},
  articleno = {294},
  booktitle = {Proceedings of the 33rd International Conference on Neural Information Processing Systems},
  numpages  = {15},
}

@inproceedings{wu-etal-2025-unveiling,
    title = "{Unveiling Environmental Impacts of Large Language Model Serving: A Functional Unit View}",
    author = "Wu, Yanran  and
      Hua, Inez  and
      Ding, Yi",
    editor = "Che, Wanxiang  and
      Nabende, Joyce  and
      Shutova, Ekaterina  and
      Pilehvar, Mohammad Taher",
    booktitle = "Proceedings of the 63rd Annual Meeting of the Association for Computational Linguistics (Volume 1: Long Papers)",
    month = jul,
    year = "2025",
    address = "Vienna, Austria",
    publisher = "Association for Computational Linguistics",
    url = "https://aclanthology.org/2025.acl-long.519/",
    doi = "10.18653/v1/2025.acl-long.519",
    pages = "10560--10576",
    ISBN = "979-8-89176-251-0"
}

\clearpage
\onecolumn
\appendix

\section{Transparency Scores}
\label{app:transparency_scores}

\begin{figure*}[htbp]
    \centering
    \includegraphics[width=\linewidth,height=0.6\textheight,keepaspectratio]{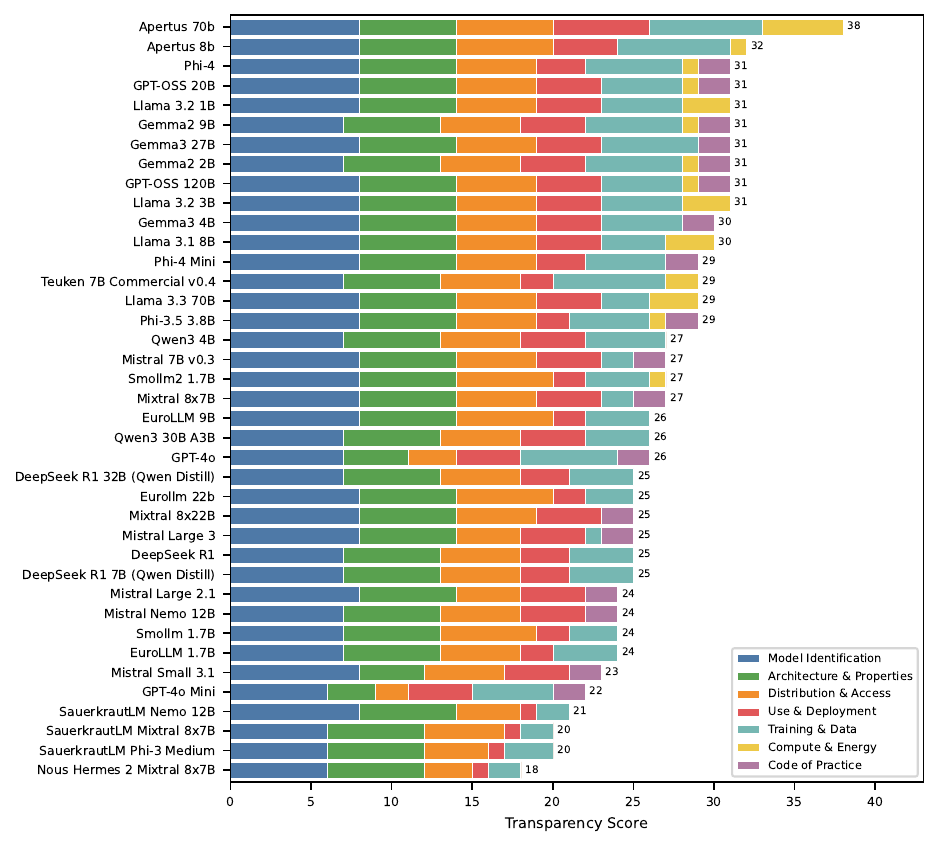}
    \caption{Transparency scores for all 39 models, sorted by domain and total score. Each segment represents the score in one of the seven transparency documentation domains. The pronounced gap between well-documented domains (Model Identification, Architecture, Distribution \& Access) and poorly-documented ones (Use \& Deployment, Training \& Data, Compute \& Energy) is visible across nearly all models.}
    \label{fig:transparency_stacked}
\end{figure*}

\end{document}